\documentclass[conference,9pt,letterpaper]{IEEEtran}
\usepackage{cite}
\usepackage{amsmath,amssymb}
\usepackage{pifont}
\usepackage{graphicx,booktabs,array,tabularx,multirow}
\usepackage{algorithm,algpseudocode}
\usepackage{stfloats}
\usepackage{placeins}
\usepackage[table]{xcolor}
\definecolor{brandred}{HTML}{9B2D3D}
\definecolor{brandblue}{HTML}{204573}
\usepackage{xspace,microtype,balance}
\usepackage[hidelinks]{hyperref}
\definecolor{tableink}{RGB}{43,44,59}
\definecolor{nativegray}{RGB}{241,243,245}
\definecolor{systemblue}{RGB}{230,231,245}
\definecolor{systemaccent}{RGB}{88,91,158}
\definecolor{cosmosaccent}{RGB}{53,90,115}
\definecolor{fastwamaccent}{RGB}{128,82,48}
\definecolor{motusaccent}{RGB}{58,96,75}
\definecolor{traccent}{RGB}{177,99,35}
\colorlet{oraccent}{systemaccent}
\definecolor{rraccent}{RGB}{48,119,101}
\definecolor{enabledgreen}{RGB}{48,119,101}
\definecolor{disabledred}{RGB}{185,61,88}
\definecolor{trbackground}{RGB}{252, 245, 236}
\definecolor{orbackground}{RGB}{244, 244, 251}
\definecolor{rrbackground}{RGB}{242, 248, 245}
\definecolor{insightbackground}{RGB}{248,248,248}
\newcommand{\tablecheck}{\textcolor{enabledgreen}{\ding{51}}}
\newcommand{\tablecross}{\textcolor{disabledred}{\ding{55}}}
\newcommand{\system}{\textsc{WAMachine}\xspace}

\newcommand{\insightbox}[2][traccent]{%
  \par\smallskip
  \begingroup
  \setlength{\fboxsep}{5pt}%
  \noindent\fcolorbox{#1}{insightbackground}{%
    \parbox{\dimexpr\linewidth-2\fboxsep-2\fboxrule\relax}{\textit{#2}}}%
  \par\endgroup
  \smallskip
}
\title{\textit{\textcolor{brandred}{WA}\textcolor{brandblue}{Machine}}: Rethinking World--Action Models\\through State Continuity}
\usepackage[most]{tcolorbox}
\usepackage{helvet}
\definecolor{FrontBoxBlue}{HTML}{F0F4F8}
\newtcolorbox{frontmatterbox}{
  enhanced,
  width=\textwidth,
  colback=FrontBoxBlue,colframe=FrontBoxBlue,
  boxrule=0pt,boxsep=0pt,arc=10pt,outer arc=10pt,
  left=7mm,right=7mm,top=7mm,bottom=7mm,
  before skip=0pt,after skip=6mm
}
\newcommand{\AIRtitleandauthors}{%
  {\raggedright\sffamily\fontsize{20}{23}\selectfont\bfseries
  \textit{\textcolor{brandred}{WA}\textcolor{brandblue}{Machine}}: Rethinking World--Action Models through State Continuity\par}
  \vspace{0.38cm}
  {\raggedright\sffamily\fontsize{10.4}{12.3}\selectfont\bfseries
  Zhinan~Liu$^{1,2,\dagger}$,
  Haozhi~Han$^{3,2,\dagger}$,
  Ruge~Zhang$^{4,2}$,
  Sa~Wang$^{4}$,
  Yunquan~Zhang$^{4}$,
  Yifeng~Chen$^{3}$,
  Teng~Ma$^{5}$,
  Zheng~Liu$^{5}$,
  Tao~Ma$^{5}$,
  Ting~Cao$^{2}$,
  Yunxin~Liu$^{2}$ and
  Kun~Li$^{2,\ddagger}$\par}
  \vspace{0.22cm}
  {\normalfont\fontsize{9.7}{11.9}\selectfont\raggedright
  $^{1}$Xiamen University\\
  $^{2}$Institute for AI Industry Research, Tsinghua University\\
  $^{3}$School of Computer Science, Peking University\\
  $^{4}$Institute of Computing Technology, Chinese Academy of Sciences\\
  $^{5}$Alibaba Group\\[2pt]
  $^{\dagger}$Equal contribution.\qquad
  $^{\ddagger}$Corresponding author.\par}
  \vspace{0.44cm}
}
\renewenvironment{abstract}
  {\begingroup\normalfont\fontsize{9.8}{12.1}\selectfont}
  {\par\endgroup}
\newcommand{\AIRcontact}{%
  \par\vspace{0.34cm}
  \noindent\begin{minipage}[b]{0.54\linewidth}
  \normalfont\fontsize{9.2}{11.2}\selectfont\raggedright
  {\sffamily\bfseries Date:} November 19, 2026\\[2pt]
  {\sffamily\bfseries Email:}
  \href{mailto:likun@air.tsinghua.edu.cn}{\textcolor{brandblue}{likun@air.tsinghua.edu.cn}}\\[2pt]
  {\sffamily\bfseries Project page:}
  \href{https://wamachine.likun.tech/}{\textcolor{brandblue}{https://wamachine.likun.tech/}}\\[2pt]
  {\sffamily\bfseries Code:}
  \href{https://github.com/RSIScience/WAMachine.git}{\textcolor{brandblue}{https://github.com/RSIScience/WAMachine.git}}\par
  \end{minipage}\hfill
  \begin{minipage}[b]{0.44\linewidth}
  \raggedleft
  \raisebox{-0.5\height}{\includegraphics[height=21.6bp,trim=0bp 0bp 263bp 0bp,clip,keepaspectratio]{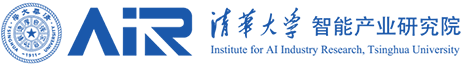}}%
  \hspace{2mm}%
  \raisebox{\dimexpr-.5\height+.396bp\relax}{\includegraphics[height=16.85bp,keepaspectratio]{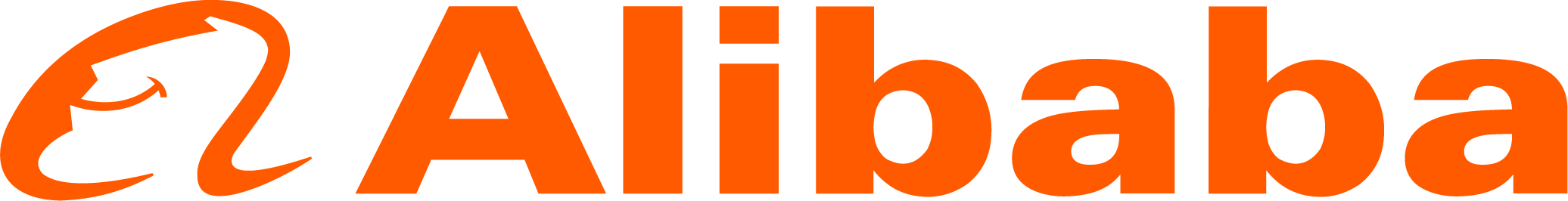}}%
  \par\vspace{2mm}
  \end{minipage}\par
}
\hypersetup{
  pdftitle={WAMachine: Rethinking World--Action Models through State Continuity},
  pdfauthor={Zhinan Liu, Haozhi Han, Ruge Zhang, Sa Wang, Yunquan Zhang, Yifeng Chen, Teng Ma, Zheng Liu, Tao Ma, Ting Cao, Yunxin Liu, Kun Li}
}

\begin{document}
\raggedbottom
\bstctlcite{IEEE:BSTcontrol}
\twocolumn[{%
\begin{frontmatterbox}
\AIRtitleandauthors
\begin{abstract}
World--Action Models (WAMs) enable future-aware control by jointly modeling actions and environment dynamics, but iterative diffusion or flow inference incurs substantial denoising latency in closed-loop control. Prior inference state offers a natural opportunity for acceleration, yet changing planning contexts, observations, and intermediate representations make direct reuse unreliable. Preserving useful computation therefore requires adapting inference state rather than reusing it as-is. To this end, we present \system, a stateful inference framework that builds on state continuity to preserve and adapt inference states across three levels of computation. Across closed-loop replans, Trajectory Remapping remaps the preceding replan's state to initialize the next. Across denoising steps, Observation Rebinding performs anticipatory inference during action execution and rebinds retained state to real observations when consistency checks pass. Across Transformer layers, Residual Rescaling rescales retained residuals and refreshes them when probe checks fail. Evaluations of three representative WAM architectures show that \system achieves 1.47--3.05$\times$ speedups in observation-to-action latency and 2.23--3.27$\times$ speedups in GPU inference time per replan, with limited success rate degradation.
\end{abstract}
\par\smallskip
\textbf{Keywords:} World--Action Models, state continuity, stateful inference, GPU runtime, embodied intelligence.
\par
\AIRcontact
\end{frontmatterbox}%
}]
\section{Introduction}
\label{sec:introduction}

World--Action Models (WAMs) jointly model robot actions and how the environment may evolve under those actions, extending visuomotor policies with future-aware prediction for closed-loop control and planning~\cite{kim2026cosmospolicy,yuan2026fastwam,bi2026motus}.
However, many WAMs generate action horizons, often together with future representations, through iterative diffusion or flow solvers, incurring substantial GPU computation and observation-to-action latency~\cite{chi2025diffusionpolicy,black2025pi0,hou2024ditpolicy}.
During closed-loop control, WAM inference involves multiple nested computation loops: an outer replanning loop repeatedly solves for new action horizons as observations arrive, an inner denoising loop iteratively refines future trajectories or representations, and the Transformer repeatedly processes intermediate states across network layers.
As a result, WAMs repeatedly recompute highly related inference states along the observation-to-action path.

\begin{figure*}[tb]
  \centering
  \includegraphics[width=\textwidth]{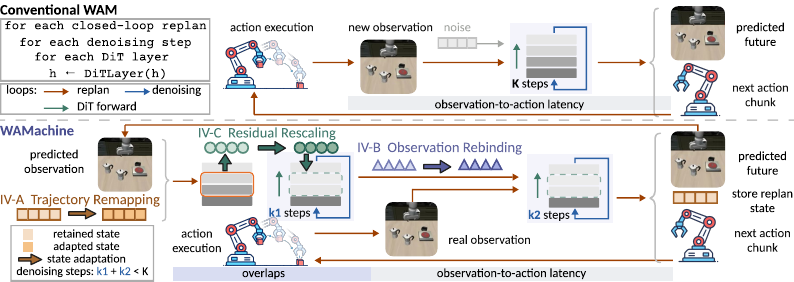}
  \caption{\textbf{Conventional WAM vs. \system.} \system adapts state across replans, denoising steps, and Transformer layers to reduce computation and latency. $K$ denotes native steps; $k_1$ and $k_2$ count steps before and after real observation arrival, with $k_1+k_2<K$ for the illustrated accepted path.}
  \label{fig:execution-comparison}
\end{figure*}

A natural direction for accelerating WAM inference is to preserve useful computation from earlier inference and adapt it as the control loop and inference context evolve.
Existing approaches explore this opportunity at different points in the inference pipeline.
RTI-DP and STEP leverage information from preceding control steps to construct better initializations for subsequent diffusion inference~\cite{duan2025rtidp,li2026step};
RTC and FutureRTC overlap action generation with physical execution to reduce computation exposed on the control path~\cite{black2025rtc,jiang2026futurertc};
and TeaCache, DiCache, and C$^3$ache reuse intermediate features or residuals across denoising steps or inference chunks to reduce repeated Transformer computation~\cite{liu2025teacache,bu2026dicache,zhao2026c3ache}.
These methods demonstrate the potential of computational reuse in iterative generative inference and robotic control.
However, existing approaches largely treat reuse as an optimization within individual inference stages, rather than considering how inference states evolve throughout the closed-loop control process.
Importantly, WAM inference exhibits \emph{\textbf{state continuity}}: intermediate states can remain informative across replans, denoising steps, and Transformer layers, even as the task context and computation evolve.
\emph{The central challenge is therefore to preserve and adapt inference states across these three levels, enabling efficient closed-loop WAM inference without redundant recomputation.}

To address this challenge, we view closed-loop WAM inference as a stateful process in which previously computed inference states can be adapted and reused as the control loop evolves.
Building on this view, we propose \textbf{\system}\footnote{\system embodies a state-machine view of WAMs, where inference evolves through the preservation and adaptation of computational states.}, a stateful inference framework that preserves and adapts inference states without additional training across three levels of computation: closed-loop replans, denoising steps, and Transformer layers.

Across closed-loop replans, \textit{Trajectory Remapping} reduces redundant trajectory generation by carrying informative replan state from the preceding replan into the next.
Since consecutive replans typically share the same task objective and exhibit gradual state evolution, the preceding solution provides a strong initialization signal despite updated observations.
It remaps these states to the new planning context, allowing WAMs to bypass unnecessary exploration from noise and accelerate closed-loop replanning.

Across denoising steps, \textit{Observation Rebinding} reduces latency exposed to the control loop by continuing partially completed inference across physical execution.
Future-aware WAMs provide predictions of upcoming observations, enabling anticipatory inference before the next observation is available.
By rebinding retained denoising states to the realized observation, it preserves computation that passes consistency checks and refreshes inference otherwise.

Across Transformer layers, \textit{Residual Rescaling} reduces repeated Transformer computation by reusing temporally coherent intermediate representations across denoising steps.
Our analysis reveals that middle-layer residuals exhibit local continuity across adjacent denoising iterations, enabling residual-level reuse with lightweight consistency checking.
It selectively rescales retained residual states and refreshes them through full computation of the middle layers when the probe checks fail, reducing DiT inference cost.


Moreover, we implement \system on existing CPU--GPU platforms, leveraging asynchronous scheduling, GPU-resident state caching, and CUDA Graphs~\cite{nvidia2026cudagraphs} to support efficient stateful inference.
We evaluate it across three representative WAM architectures on LIBERO~\cite{liu2023libero} and RoboTwin 2.0~\cite{chen2026robotwin2}, analyze the contribution of each mechanism, and study success--latency trade-offs under different inference budgets.

Our main contributions are as follows:
\begin{itemize}



    \item We reveal state continuity as a new acceleration opportunity in WAM inference and introduce stateful inference for preserving and adapting evolving inference states.

    \item We propose \system, a stateful inference framework that preserves and adapts inference states across replanning, denoising, and Transformer execution through trajectory remapping, observation rebinding, and residual rescaling.

    \item \system achieves 1.47--3.05$\times$ speedups in observation-to-action latency and 2.23--3.27$\times$ speedups in GPU inference time per replan for three representative WAMs on LIBERO and RoboTwin 2.0, with limited success rate degradation.
    
\end{itemize}

Experimental results show that, compared with native WAM inference, \system achieves 1.47--3.05$\times$ speedups in observation-to-action latency and 2.23--3.27$\times$ speedups in GPU inference time per replan, with limited success rate degradation in large-sample evaluations.
Ablation studies on Fast-WAM-IDM~\cite{yuan2026fastwam} further show that the complete system achieves the lowest observation-to-action latency among the tested variants.

\section{Background}
\label{sec:background}

\subsection{World--Action Models}
\textbf{Model Architecture.} World--Action Models (WAMs) jointly model robot actions and future environment evolution, enabling future-aware control. Cosmos Policy builds on a pretrained video diffusion backbone to generate action trajectories, future observations, and value estimates~\cite{kim2026cosmospolicy}. Motus brings understanding, video generation, and action prediction together through a Mixture-of-Transformers architecture~\cite{liang2025mot,bi2026motus}, while $\tau_0$-WM couples video--action prediction with action-conditioned future simulation and task-progress evaluation~\cite{zhou2026tau0wm}.

\textbf{Action Chunks and Replanning.} At replan $t$, the observation $o_t$, robot state $s_t$ when available, and task instruction $g$ form the conditioning context $c_t=(o_t,s_t,g)$ to generate an \emph{action chunk} of $H$ future actions, $A_t=(a_{t,0},a_{t,1},\ldots,a_{t,H-1})$.
Across closed-loop replans, the robot executes a prefix of $A_t$ and generates the next chunk from an updated observation, forming the outer \emph{replanning loop}~\cite{zhao2023act,chi2025diffusionpolicy}.
\looseness=-1

\textbf{Diffusion and Flow Solvers.} Across denoising steps within each replan, a diffusion or flow solver~\cite{ho2020ddpm,karras2022edm,lipman2023flowmatching} progressively refines a noisy sample through $K$ updates \mbox{$x_t^{i+1}=\mathcal S_i\!\left(x_t^i,f_\theta(x_t^i,i;c_t)\right)$} for $0\le i<K$, where $x_t^i$ is the intermediate sample, the neural network $f_\theta$ predicts a denoising quantity or flow velocity, and $\mathcal S_i$ denotes the solver update~\cite{chi2025diffusionpolicy,black2025pi0}. Depending on the architecture, the sample may represent noisy actions, future-state latents, or their joint representation. These repeated solver updates form the \emph{denoising loop}.

\textbf{Transformer Execution.} At each denoising step, $f_\theta$ typically evaluates a diffusion Transformer (DiT) with $L$ blocks~\cite{peebles2023dit}. For a fixed replan, computation proceeds across Transformer~\cite{vaswani2017attention} layers through residual updates~\cite{he2016resnet} $h_{\ell+1}(i)=h_\ell(i)+\mathcal F_\ell(h_\ell(i);c_t,i)$, where $h_\ell(i)$ denotes the \emph{hidden state} after $\ell$ blocks, and $\mathcal F_\ell$ denotes the block's net residual transformation, including attention and feed-forward computation. The cumulative residual over a span of layers is the difference between its output and input hidden states. These layer-wise updates form the innermost \emph{Transformer-layer execution loop}, nested within the denoising and replanning loops.

\subsection{Inference Acceleration Approaches}
\textbf{Existing Approaches.} Prior work optimizes model architectures and inference. Fast-WAM skips test-time future generation, while Faster-WAM docks a lightweight action head onto a video backbone~\cite{yuan2026fastwam,ma2026fasterwam}. RTI-DP warm-starts from prior solutions, while STEP predicts consistent warm starts~\cite{duan2025rtidp,li2026step}. RTC overlaps chunk generation and execution via inpainting; FutureRTC predicts execution-time observations and states~\cite{black2025rtc,jiang2026futurertc}. TeaCache adaptively reuses denoising outputs, DiCache uses shallow probes for cache scheduling and alignment, and C$^3$ache reuses residuals across chunks at matching denoising steps~\cite{liu2025teacache,bu2026dicache,zhao2026c3ache}.
\looseness=-1

\textbf{Distinction from Prior Work.} Compared with RTI-DP and STEP, Trajectory Remapping reconstructs stage-aligned denoising states from retained endpoints and normalized directions. Beyond RTC and FutureRTC's asynchronous chunking, Observation Rebinding conditionally continues latents and solver history from predicted to real conditions. Unlike DiCache's cache alignment and C$^3$ache's cross-chunk reuse, Residual Rescaling adapts cumulative middle-layer residuals with a probe-fitted scale, gating stepwise reuse by directional agreement and fitting error. Together, they adapt states across replanning, observation updates, and denoising.
\looseness=-1

\section{Motivation}
\label{sec:stateful}
In this section, we progressively analyze state continuity across closed-loop replans, denoising steps, and Transformer layers, identify reuse challenges under evolving conditions, and derive insights for preserving and adapting inference states.
\looseness=-1

\subsection{Replan State Continuity}
\textit{Motivation 1: Native WAM inference restarts denoising from noise at each replan, incurring repeated computation. In contrast, consecutive replans share the same task objective under gradually evolving states, offering continuity across replan states.}
\looseness=-1

\begin{figure}[!t]
\centering
\includegraphics[width=\columnwidth]{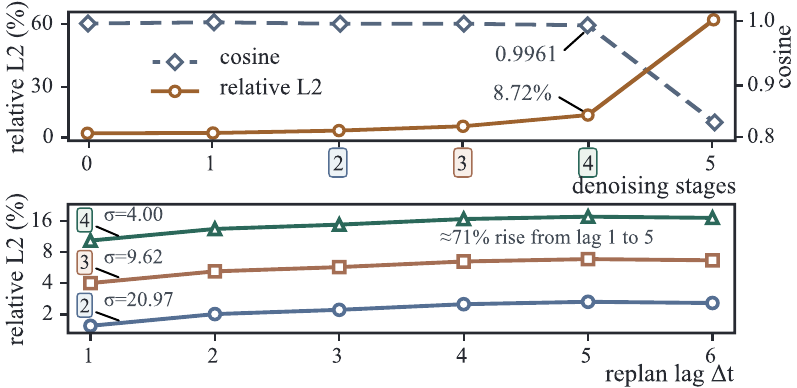}
\caption{\textbf{Replan state continuity.}}
\label{fig:continuity-replan}
\end{figure}

\textbf{Stage-Dependent Continuity.} We first examine trajectory latents between consecutive replans of Native Cosmos Policy across denoising stages. In Fig.~\ref{fig:continuity-replan} (top), the latents exhibit high similarity at early denoising stages, with a cosine similarity of $0.9961$ and relative L2 distance of $8.72\%$ at stage 4. At the final stage, relative L2 distance rises above $60\%$, indicating that continuity between replan states varies across denoising stages.

\textbf{Temporal Locality.} We further investigate how replan lag affects state continuity. Within each episode, we compare native action latents separated by $\Delta t=1$--$6$ replans at stages 2, 3, and 4, corresponding to noise levels $\sigma=20.97, 9.62,$ and $4.00$~\cite{karras2022edm}, where larger $\sigma$ denotes earlier, noisier trajectories. In Fig.~\ref{fig:continuity-replan} (bottom), mean relative L2 distance increases by approximately 71\% from lag 1 to lag 5 across all three noise levels.

\insightbox[traccent]{Insight 1: Replan states exhibit local, stage-dependent continuity, suggesting that recently computed replan states can be adapted to initialize subsequent replans.}

\subsection{Denoising State Continuity}
\textit{Motivation 2: Future-aware WAMs predict upcoming observations, creating an opportunity for denoising during action execution. Directly using these predictions for denoising can lead to substantial action errors when they deviate from real observations.}
\looseness=-1

\begin{figure}[!t]
\centering
\includegraphics[width=\columnwidth]{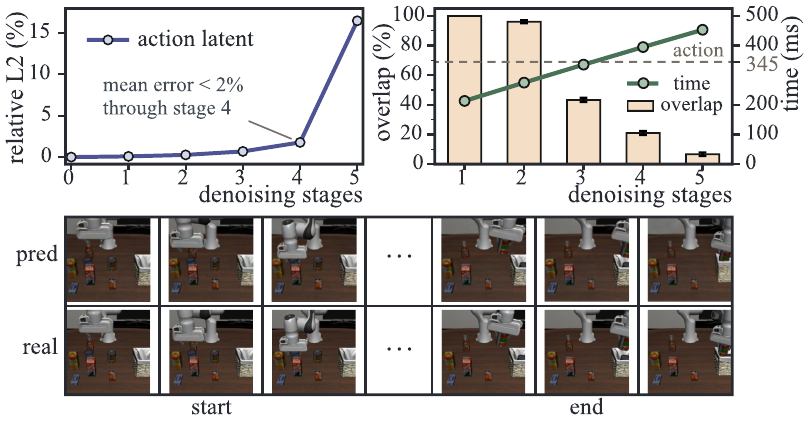}
\caption{\textbf{Denoising state continuity.}}
\label{fig:continuity-denoising}
\par\vspace{7pt}
\centering
\includegraphics[width=\columnwidth]{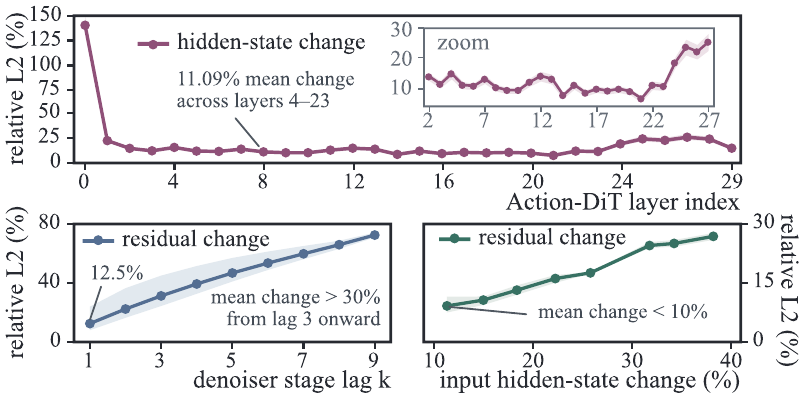}
\caption{\textbf{Layer state continuity.}}
\label{fig:continuity-layer}
\end{figure}

\textbf{Cross-Condition Continuity.} Even with initialized replan states, waiting for real observations leaves the action execution window unused for denoising. In Fig.~\ref{fig:continuity-denoising} (bottom), predicted observations closely resemble real counterparts despite robot pose and object differences. Performing independent native denoising under both conditions from identical initial noise, we observe in Fig.~\ref{fig:continuity-denoising} (top left) that intermediate action latents maintain a mean relative L2 difference below 2\% through stage 4 before rising sharply at the final stage, allowing early denoising computation to be preserved across observation changes.

\begin{figure*}[!t]
\centering
\includegraphics[width=\textwidth]{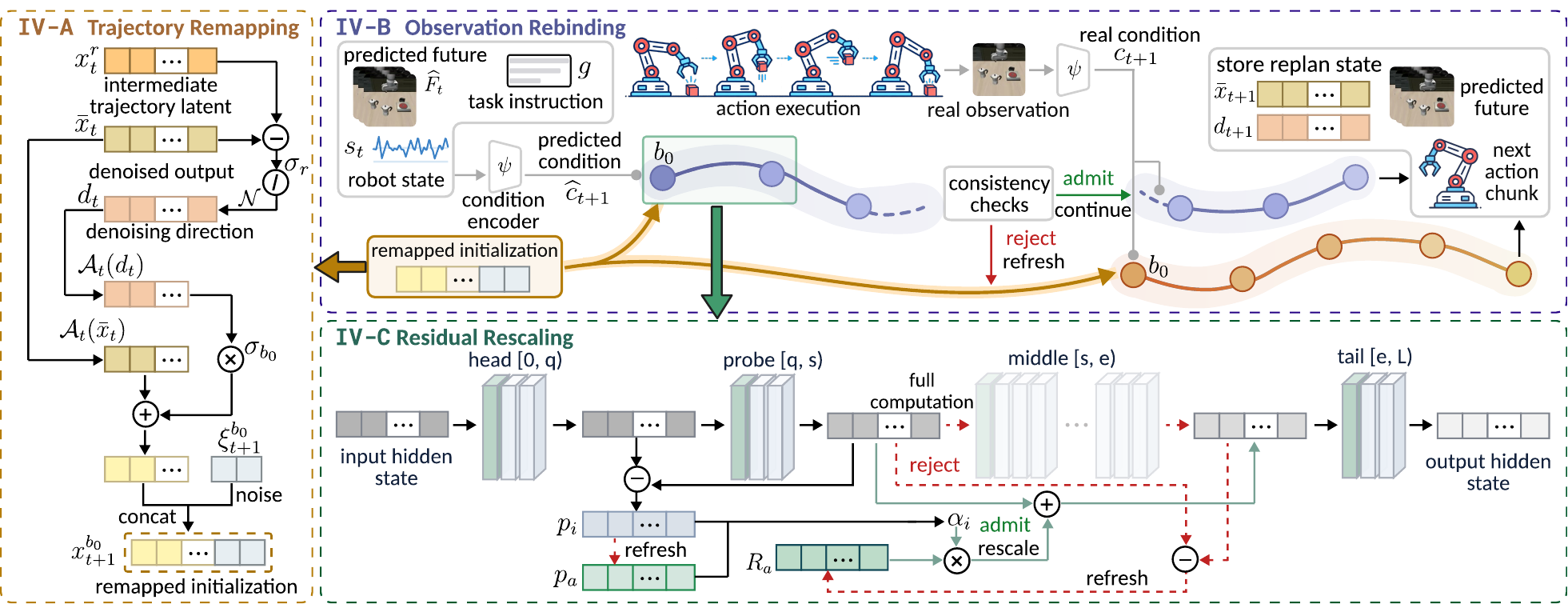}
\caption{\textbf{The \system framework.}}
\label{fig:framework}
\end{figure*}

\textbf{Execution-Window Limit.} We next examine whether these early denoising stages fit within the action execution window. As shown in Fig.~\ref{fig:continuity-denoising} (top right), prefix latency grows with the number of denoising steps and crosses the 345\,ms window after stage 3, reducing the fraction of completed prefixes. Earlier stages therefore offer greater overlap opportunities, while the execution window limits how far denoising can advance before observation arrival.

\insightbox[oraccent]{Insight 2: Early denoising states exhibit continuity across observation conditions. The limited execution window motivates a bounded prefix during action execution and adaptation of retained state to the real observation for continuation.}

\subsection{Layer State Continuity}
\textit{Motivation 3: A Transformer progressively refines the input through many layers, incurring repeated computation across steps. This raises the question of whether intermediate transformations remain reusable as the denoising input evolves.}
\looseness=-1

\textbf{Layer-Wise Evolution.} We first examine how hidden states evolve across Transformer layers in Fast-WAM-IDM's Action-DiT. As shown in Fig.~\ref{fig:continuity-layer} (top), relative L2 changes between adjacent hidden states are large in the earliest layers but much smaller in the middle layers. To characterize the middle layers, we measure their cumulative residual, defined as the difference between the output and input hidden states.

\textbf{Residual Continuity.} We next compare these residuals across denoising steps using relative L2 differences. In Fig.~\ref{fig:continuity-layer} (bottom left), residual change is 12.5\% between adjacent steps and increases with step lag, showing that continuity weakens over longer intervals. To examine how the current input relates to this continuity, we further compare residual changes with hidden-state changes at the middle-layer input. In Fig.~\ref{fig:continuity-layer} (bottom right), residual change rises with input change, remaining below 10\% on average for small input changes.

\insightbox[rraccent]{Insight 3: Middle-layer residuals exhibit local continuity across denoising steps, motivating reuse of recent residuals while accounting for changes in the current input.}

\section{WAMachine Framework}
\label{sec:method}
The analysis and insights motivate us to preserve and adapt inference states.
In this section, we detail how \system coordinates Trajectory Remapping (TR), Observation Rebinding (OR), and Residual Rescaling (RR) to address reuse challenges across closed-loop replans, denoising steps, and Transformer layers, as shown in Fig.~\ref{fig:framework}.
\looseness=-1

\begingroup
\setlength{\abovedisplayskip}{4pt plus 1pt minus 1pt}
\setlength{\belowdisplayskip}{4pt plus 1pt minus 1pt}
\setlength{\abovedisplayshortskip}{0pt plus 1pt}
\setlength{\belowdisplayshortskip}{3pt plus 1pt minus 1pt}

\subsection{Trajectory Remapping across Closed-Loop Replans}
\label{sec:tr}
From Insight 1, we propose TR to reduce redundant trajectory generation. As illustrated on the left of Fig.~\ref{fig:framework}, it remaps prior replan state to initialize the next replan with fewer denoising steps.

Specifically, TR retains each branch's final denoised output $\bar x_t$ and normalized direction \mbox{$d_t=\mathcal N((x_t^r-\bar x_t)/\sigma_r)$}, computed from an intermediate latent $x_t^r$ at native stage $r$ with noise level $\sigma_r>0$, where \mbox{$\mathcal N(v)=v/\sqrt{\operatorname{mean}(v^2)}$}.
TR applies the same remapping operator $\mathcal A_t$ to $\bar x_t$ and $d_t$. At entry stage $b_0$ with noise level $\sigma_{b_0}$, it initializes target slots as
\begin{equation}
 x_{t+1}^{b_0}[j]=
 \begin{cases}
 \mathcal A_t(\bar x_t)[j]+\sigma_{b_0}\mathcal A_t(d_t)[j],&j\in\mathcal M_t,\\
 \xi_{t+1}^{b_0}[j],&j\notin\mathcal M_t,
 \end{cases}
 \label{eq:transport}
\end{equation}
where $\mathcal M_t$ contains target slots with retained sources, and $\xi_{t+1}^{b_0}$ supplies fresh noise elsewhere. The first replan uses fresh noise throughout. Starting at $b_0$ skips earlier updates, leaving fewer denoising steps for refinement.
\looseness=-1

Cosmos Policy retains joint action and future-video states, while Motus retains action states, both without shifting. Fast-WAM-IDM maps action slots $j+10\mapsto j$ after ten executed actions, retains video indices, and fills the final ten action slots with fresh noise.
\looseness=-1

\subsection{Observation Rebinding across Denoising Steps}
\label{sec:or}
From Insight 2, we propose OR to reduce observation-to-action latency. As shown in the upper right of Fig.~\ref{fig:framework}, it advances denoising under predicted observations during action execution, then rebinds retained state to the real observation when consistency checks pass.
\looseness=-1

Specifically, during physical action execution, OR forms a predicted condition $\widehat c_{t+1}=\psi(\widehat F_t,s_t,g)$ from the predicted future $\widehat F_t$, robot state $s_t$, and instruction $g$. Starting from TR's initialization, it advances to a predefined rebinding stage, retaining the latent and native solver history without issuing actions.

When the real observation arrives, OR compares predicted camera images and robot states $\widehat u_m$ with their observed counterparts $u_m$ before encoding. It aggregates normalized errors \mbox{$e_m=\sqrt{\operatorname{mean}[((\widehat u_m-u_m)/\delta_m)^2]}$} into a weighted root-mean-square consistency score \mbox{$S=\sqrt{(\sum_m w_m e_m^2)/(\sum_m w_m)}$}, where $\delta_m$ and $w_m$ denote the corresponding normalization scale and weight. After the encoder $\psi$ constructs the real condition $c_{t+1}$, OR replaces $\widehat c_{t+1}$ with $c_{t+1}$ and continues from the retained state if $S<\tau$ and the consecutive-acceptance counter $\kappa<\kappa_{\max}$; otherwise, it restarts from TR's initialization under $c_{t+1}$. Each accepted prefix increments $\kappa$, and a refresh resets it. Reaching the cap forces a real-conditioned refresh at the next replan and suppresses its prefix.

\begin{figure}[!t]
\centering
\includegraphics[width=\columnwidth]{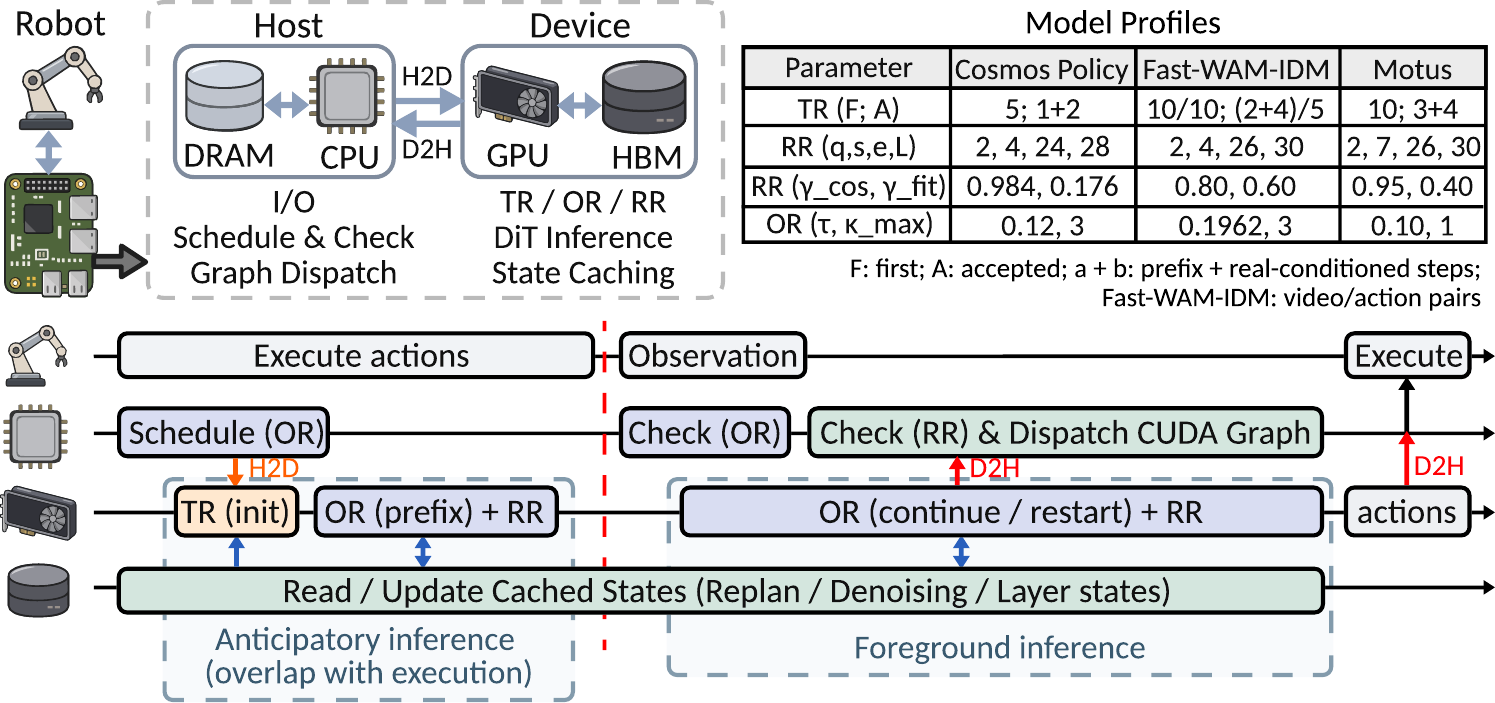}
\caption{\textbf{The \system implementation.}}
\label{fig:implementation}
\end{figure}

\begin{algorithm}[!b]
\caption{Stateful Inference}
\label{alg:stateful-inference}
\small
\algrenewcommand{\algorithmicindent}{1em}
\algtext*{EndIf}
\makeatletter
\newcommand{\algloopshade}[1]{%
  \smash{\rlap{%
    \hspace*{-2pt}%
    \textcolor{#1}{%
      \rule[-.22\baselineskip]{\dimexpr\linewidth-\ALG@thistlm+2pt\relax}{.95\baselineskip}}%
  }}%
}
\makeatother
\algrenewtext{For}[1]{\algloopshade{trbackground}\algorithmicfor\ #1\ \algorithmicdo}
\algrenewtext{EndFor}{\algloopshade{trbackground}\algorithmicend\ \algorithmicfor}
\def\algloopcolor{orbackground}
\algrenewtext{While}[1]{\algloopshade{\algloopcolor}\algorithmicwhile\ #1\ \algorithmicdo}
\algrenewtext{EndWhile}{\algloopshade{\algloopcolor}\algorithmicend\ \algorithmicwhile}
\begin{algorithmic}[1]
\Require Instruction $g$; model profile $\Pi$ with $\tau,\kappa_{\max}$
\State $A,P,\widehat F,C\gets\varnothing$; $\kappa\gets0$
\For{each closed-loop replan $t$} \Comment{Replanning loop: TR}
  \State Dispatch $A$ asynchronously if $A\ne\varnothing$
  \State $v\gets(\widehat F\ne\varnothing\ \land\ \kappa<\kappa_{\max})$; select schedule from $\Pi$
  \State $(I,b)\gets(\text{native noise},0)$ if $P=\varnothing$, else $(\operatorname{TR}(P),b_0)$
  \State Receive $(O,s_t)$ if $\neg v$; encode $c$ from $\widehat F$ if $v$, else $O$
  \State $Y\gets(I,b,\varnothing)$; reset $\kappa$ if $\neg v$
  \While{denoising is unfinished} \Comment{Denoising loop: OR}
    \If{$v$ and the rebinding stage is reached}
      \State Receive $(O,s_t)$; $\mathrm{accept}\gets[S(\widehat F,O)<\tau]$
      \State Keep $Y$ if accepted; otherwise $Y\gets(I,b,\varnothing)$
      \State $\kappa\gets\kappa+1$ if accepted, else $0$; select schedule
      \State $c\gets\psi(O,s_t,g)$; $v\gets\mathrm{false}$
    \EndIf
    \State $(x,i,\mathcal H)\gets Y$; $h_0\gets\operatorname{Embed}(x;c,i)$; $\ell\gets0$
    \def\algloopcolor{rrbackground}
    \While{$\ell<L$} \Comment{Transformer-layer loop: RR}
      \If{$\ell=s$, $C\ne\varnothing$, and RR$(h_s-h_q,C)$ yields valid $\alpha_i$}
        \State $(p_a,R_a)\gets C$; $h_e\gets h_s+\alpha_iR_a$; $\ell\gets e$; \textbf{continue}
      \EndIf
      \State $h_{\ell+1}\gets h_\ell+\mathcal F_\ell(h_\ell;c,i)$; $\ell\gets\ell+1$
      \State \textbf{if} $\ell=e$ \textbf{then} $C\gets(h_s-h_q,h_e-h_s)$
    \EndWhile
    \State Save $x^r\gets x$ if $i=r$; $Y\gets\operatorname{NativeStep}(Y,\operatorname{Project}(h_L))$
    \def\algloopcolor{orbackground}
  \EndWhile
  \State Decode $A,\widehat F$ from $Y$; build $P$ from $x^r$ and final latent by TR
\EndFor
\end{algorithmic}
\end{algorithm}

\subsection{Residual Rescaling across DiT Layers}
\label{sec:rr}
From Insight 3, we propose RR to reduce repeated Transformer computation. As shown in the lower right of Fig.~\ref{fig:framework}, it rescales retained residuals using a shallow probe, skipping middle layers when checks pass and recomputing them otherwise.
\looseness=-1

Specifically, partition an $L$-layer diffusion Transformer~\cite{peebles2023dit} into a head $[0,q)$, shallow probe $[q,s)$, middle $[s,e)$, and tail $[e,L)$. After full evaluation at step $a$, RR retains the probe residual \mbox{$p_a=h_s(a)-h_q(a)$} and cumulative middle-layer residual \mbox{$R_a=h_e(a)-h_s(a)$}.
At step $i$, RR evaluates the head and probe on the current input to obtain \mbox{$p_i=h_s(i)-h_q(i)$}. It then fits $\alpha_i$ so that $\alpha_i p_a$ approximates $p_i$, using the probe comparison to adapt $R_a$:
\begin{equation}
 \alpha_i=\operatorname{clamp}\!\left(
 \frac{\operatorname{mean}(p_i\odot p_a)}
 {\max(\operatorname{mean}(p_a\odot p_a),\epsilon)},
 \alpha_{\min},\alpha_{\max}\right).
 \label{eq:alpha}
\end{equation}
It then requires sufficient directional agreement \mbox{$\cos(p_i,p_a)\ge\gamma_{\cos}$} and a small relative fitting error \mbox{$\|p_i-\alpha_i p_a\|_2/\|p_i\|_2\le\gamma_{\mathrm{fit}}$}.
If both checks pass, RR applies the fitted scale to $R_a$ and adds the rescaled residual to the current $h_s(i)$, skipping the middle layers:
\begin{equation}
 \widetilde h_e(i)=h_s(i)+\alpha_iR_a.
 \label{eq:inject}
\end{equation}
Otherwise, RR computes the middle layers from the already available $h_s(i)$ and refreshes $(p_a,R_a)$ with exact residuals. In either case, inference then proceeds through the tail. After observation rebinding, RR applies the same checks under the real condition.
\par
\endgroup

\section{System Implementation}
\label{sec:runtime}
This section describes how \system combines asynchronous scheduling, GPU-resident caching, and CUDA Graphs on existing CPU--GPU platforms to support efficient stateful inference.

\subsection{CPU--GPU Execution and State Caching}
On the host side, the CPU handles observation and action I/O, asynchronous scheduling, consistency checks, and graph dispatch. On the device side, the GPU executes DiT inference and state adaptation across TR, OR, and RR. HBM caches TR replan states, OR intermediate latents and solver history, and RR probe and middle-layer residuals, allowing GPU kernels to read and update states directly.

\system captures CUDA Graphs~\cite{nvidia2026cudagraphs} for full computation, head-and-probe execution, accepted residual reuse, and post-rejection continuation. Inputs are staged in fixed-address buffers via device-to-device (D2D) copies as needed. The CPU uses compact device-to-host (D2H) probe statistics to dispatch reuse or continuation graphs.

\begingroup
\edef\SavedFigureCounter{\number\value{figure}}
\setcounter{figure}{7}
\begin{figure}[!b]
\centering
\includegraphics[width=\columnwidth]{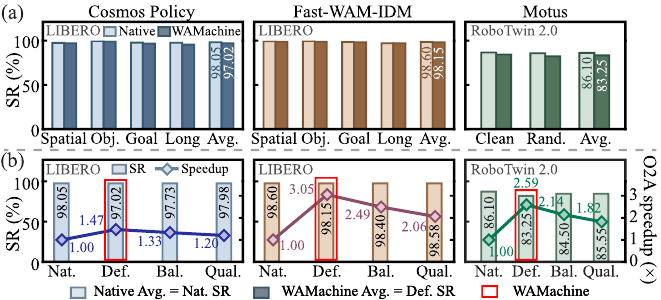}
\caption{\textbf{Task success and success--latency trade-offs.}}
\label{fig:success-tradeoff}
\end{figure}
\setcounter{figure}{\SavedFigureCounter}
\endgroup

\subsection{Overall Workflow of \textit{WAMachine}}
In Fig.~\ref{fig:implementation}, the CPU schedules OR's anticipatory prefix during action execution using TR initialization and RR; scheduled refreshes bypass it. On real observation arrival, the runtime waits for any unfinished prefix to reach the rebinding stage, then checks consistency to continue from retained state or restart under the real condition. Decoded actions return to the CPU; states remain GPU-resident. Algorithm~\ref{alg:stateful-inference} coordinates three nested loops with replan state $P=(\bar{x},d)$ for the final denoised output and normalized direction, denoising state $Y=(x,i,\mathcal H)$ for the latent, denoising stage, and solver history, and layer state $C=(p_a,R_a)$ for the probe and cumulative middle-layer residuals per DiT branch.
\looseness=-1

\subsection{Model Profiles and Configuration}
Fig.~\ref{fig:implementation} summarizes the default denoising schedules, layer partitions, and thresholds. Across all three models, RR is enabled on every execution path, with $\alpha_i$ clipped to $[1,1.25]$ and $\epsilon=10^{-8}$. Profile parameters are calibrated once on 100 diagnostic samples with disjoint seeds and fixed across tasks.

\section{Experiments}
\label{sec:evaluation}
\begin{table*}[t]
\centering
\color{tableink}
\arrayrulecolor{tableink}
\setlength{\abovecaptionskip}{2pt}
\caption{Comparison with Acceleration Baselines. Small Text Denotes Speedup over Native}
\label{tab:efficiency}
\small
\renewcommand{\arraystretch}{0.9}
\setlength{\tabcolsep}{2pt}
\newcommand{\timecell}[3][]{\mbox{#1{#2}\,{\scriptsize #1{#3\texttimes}}}}
\begin{tabularx}{\textwidth}{@{}l*{3}{*{2}{>{\hsize=1.15\hsize\linewidth=\hsize\centering\arraybackslash}X}>{\hsize=0.7\hsize\linewidth=\hsize\centering\arraybackslash}X}@{}}
\toprule
\multirow{2}{*}{Method} & \multicolumn{3}{c}{\textcolor{cosmosaccent}{\textbf{Cosmos Policy (LIBERO)}}} & \multicolumn{3}{c}{\textcolor{fastwamaccent}{\textbf{Fast-WAM-IDM (LIBERO)}}} & \multicolumn{3}{c}{\textcolor{motusaccent}{\textbf{Motus (RoboTwin 2.0 Clean)}}}\\
\cmidrule(lr){2-4}\cmidrule(lr){5-7}\cmidrule(l){8-10}
& GPU (ms) $\downarrow$ & O2A (ms) $\downarrow$ & SR (\%) $\uparrow$ & GPU (ms) $\downarrow$ & O2A (ms) $\downarrow$ & SR (\%) $\uparrow$ & GPU (ms) $\downarrow$ & O2A (ms) $\downarrow$ & SR (\%) $\uparrow$\\
\midrule
\rowcolor{nativegray}
Native & \timecell{284.29}{1.00} & \timecell{636.77}{1.00} & \textbf{98.0} & \timecell{367.49}{1.00} & \timecell{416.64}{1.00} & \underline{98.0} & \timecell{1455.69}{1.00} & \timecell{2343.14}{1.00} & \underline{84.0}\\
RTI-DP & \timecell[\textbf]{59.14}{4.81} & \timecell[\textbf]{399.15}{1.60} & 68.5 & \timecell{269.27}{1.36} & \timecell{306.64}{1.36} & 63.0 & \timecell[\textbf]{153.07}{9.51} & \timecell[\textbf]{688.37}{3.40} & 8.5\\
RTC & \timecell{1024.53}{0.28} & \timecell{2203.99}{0.29} & \textbf{98.0} & \timecell{729.27}{0.50} & \timecell{1959.68}{0.21} & 82.0 & \timecell{5639.22}{0.26} & \timecell{12683.43}{0.18} & 82.0\\
VLA-Cache & \timecell{461.48}{0.62} & \timecell{813.54}{0.78} & \textbf{98.0} & \timecell{636.89}{0.58} & \timecell{714.84}{0.58} & 96.5 & \timecell{1525.04}{0.95} & \timecell{2539.52}{0.92} & \textbf{86.5}\\
BAC & \timecell{205.53}{1.38} & \timecell{564.98}{1.13} & \underline{97.5} & \timecell{276.12}{1.33} & \timecell{324.39}{1.28} & \textbf{99.0} & \timecell{973.63}{1.50} & \timecell{1894.10}{1.24} & 82.0\\
DiCache & \timecell{167.32}{1.70} & \timecell{528.22}{1.21} & 95.0 & \timecell{237.87}{1.54} & \timecell{286.23}{1.46} & 95.5 & \timecell{1010.67}{1.44} & \timecell{1898.56}{1.23} & 82.0\\
C$^3$ache & \timecell{139.67}{2.04} & \timecell{495.98}{1.28} & 96.0 & \timecell[\underline]{233.21}{1.58} & \timecell[\underline]{281.07}{1.48} & 96.5 & \timecell{856.78}{1.70} & \timecell{1754.34}{1.34} & 82.0\\
\rowcolor{systemblue}
\textcolor{systemaccent}{\system} & \timecell[\underline]{120.29}{2.36} & \timecell[\underline]{434.64}{1.47} & 96.0 & \timecell[\textbf]{112.52}{3.27} & \timecell[\textbf]{136.42}{3.05} & 97.0 & \timecell[\underline]{652.26}{2.23} & \timecell[\underline]{906.38}{2.59} & 82.0\\
\bottomrule
\end{tabularx}
\arrayrulecolor{black}
\end{table*}

\begin{figure*}[!t]
\centering
\includegraphics[width=\textwidth]{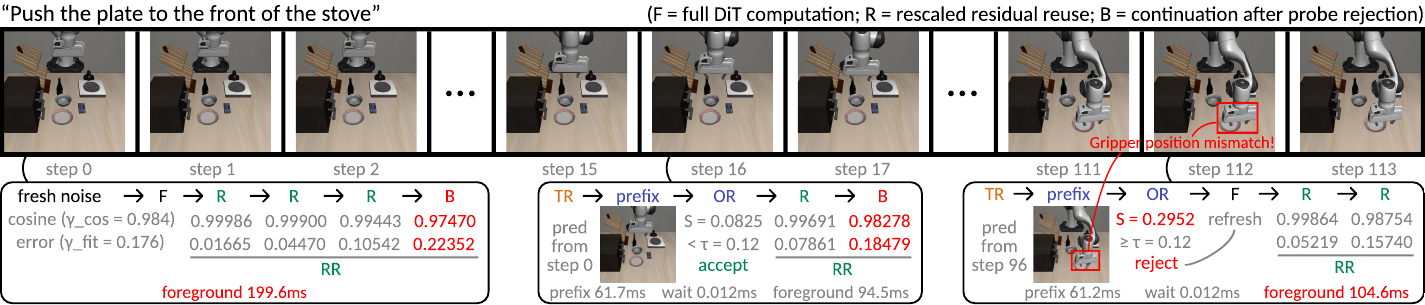}
\caption{\textbf{An execution example of \system.}}
\label{fig:execution-example}
\end{figure*}

\stepcounter{figure}

\subsection{Setup}
We evaluate \system on three WAMs~\cite{kim2026cosmospolicy,yuan2026fastwam,bi2026motus} with default profiles in Fig.~\ref{fig:implementation}: Cosmos Policy and Fast-WAM-IDM, which retains test-time future generation, across LIBERO's Spatial, Object, Goal, and Long suites of 10 tasks each~\cite{liu2023libero}, and Motus across 50 RoboTwin 2.0 tasks under Clean and Randomized settings~\cite{chen2026robotwin2}. Success evaluation uses 6,000 episodes per LIBERO model and 2,000 on RoboTwin; latency uses matched 200-episode subsets. Table~\ref{tab:efficiency} baselines include Native, RTI-DP~\cite{duan2025rtidp}, RTC~\cite{black2025rtc}, VLA-Cache~\cite{xu2025vlacache}, BAC~\cite{ji2026bac}, DiCache~\cite{bu2026dicache}, and C$^3$ache~\cite{zhao2026c3ache}, adapted per architecture and evaluated without CUDA Graphs. \system enables them as part of its runtime optimization. On one NVIDIA A100 80GB GPU per worker, we report success rate (SR), CUDA-event GPU time for DiT including anticipatory and refreshed work, and wall-clock observation-to-action (O2A) latency from input availability to decoded-action readiness.
\looseness=-1

\subsection{Evaluation Results}
\textbf{Task Success.} Fig.~\ref{fig:success-tradeoff}(a) reports large-sample task success.

On LIBERO, \system achieves average SRs of $97.02\%$ and $98.15\%$ on Cosmos Policy and Fast-WAM-IDM, compared with Native's $98.05\%$ and $98.60\%$, respectively. On RoboTwin 2.0, Motus with \system achieves $84.30\%$ SR in Clean and $82.20\%$ in Randomized. These large-sample results complement the inference speedups reported in Table~\ref{tab:efficiency}.

\begin{table}[t]
\centering
\color{tableink}
\arrayrulecolor{tableink}
\setlength{\abovecaptionskip}{2pt}
\caption{Component Ablation on Fast-WAM-IDM}
\label{tab:ablation}
\small
\renewcommand{\arraystretch}{0.9}
\setlength{\tabcolsep}{2pt}
\begin{tabularx}{\columnwidth}{@{}lccc*{3}{>{\centering\arraybackslash}X}@{}}
\toprule
Variant & \textcolor{traccent}{TR} & \textcolor{systemaccent}{OR} & \textcolor{rraccent}{RR} & GPU (ms) $\downarrow$ & O2A (ms) $\downarrow$ & SR (\%) $\uparrow$\\
\midrule
\rowcolor{nativegray}
Native & \tablecross & \tablecross & \tablecross & 415.19 & 470.96 & 97.5\\
\rowcolor{systemblue}
\textcolor{systemaccent}{\system} & \tablecheck & \tablecheck & \tablecheck & 114.75 & 144.75 & 98.0\\
w/o TR & \tablecross & \tablecheck & \tablecheck & 188.02 & 177.20 & 97.0\\
w/o OR & \tablecheck & \tablecross & \tablecheck & 108.45 & 165.62 & 95.5\\
w/o RR & \tablecheck & \tablecheck & \tablecross & 188.99 & 205.45 & 99.0\\
\bottomrule
\end{tabularx}
\arrayrulecolor{black}
\end{table}

\begin{figure}[!tb]
\centering
\includegraphics[width=\columnwidth]{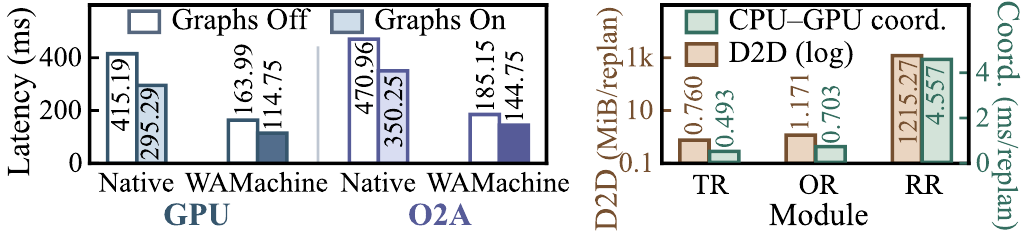}
\caption{\textbf{Runtime analysis on Fast-WAM-IDM.}}
\label{fig:hardware-runtime}
\end{figure}
\textbf{Acceleration and Baseline Comparison.} Table~\ref{tab:efficiency} shows consistent acceleration across all three architectures. Relative to Native, \system achieves $1.47\times$, $3.05\times$, and $2.59\times$ O2A speedups and $2.36\times$, $3.27\times$, and $2.23\times$ GPU speedups on Cosmos Policy, Fast-WAM-IDM, and Motus, respectively. It also achieves lower GPU and O2A latency than BAC, DiCache, and C$^3$ache on every model. Compared with C$^3$ache, the fastest of these three baselines, O2A latency is further reduced by $12.4\%$, $51.5\%$, and $48.3\%$, respectively. RTC and VLA-Cache increase latency across all models, while RTI-DP's single-step warm start yields lower GPU time on Cosmos and Motus with subset SRs of only $68.5\%$ and $8.5\%$.

On Fast-WAM-IDM, TR reduces denoising steps, while RR skips $45.80\%$ of full-depth evaluations. OR additionally advances denoising during action execution, with accepted prefixes covering $71.40\%$ of noninitial replans. Since GPU time includes anticipatory work, the GPU speedup reflects reduced DiT computation.

\textbf{Example Presentation.} Fig.~\ref{fig:execution-example} illustrates a LIBERO-Goal rollout, starting from fresh noise with $199.6$\,ms of foreground inference. RR reuses rescaled residuals after full DiT computation and recomputes middle layers upon probe rejection. During action execution, TR initializes OR's anticipatory prefix under predicted observations. At step 16, OR accepts the $61.7$\,ms prefix with $S=0.0825<\tau=0.12$; real-conditioned continuation takes $94.5$\,ms of foreground inference. At step 112, a gripper-position mismatch with the step-96 prediction yields $S=0.2952\ge\tau$. OR rejects the $61.2$\,ms prefix; restarting from TR's initialization under the real condition takes $104.6$\,ms of foreground inference. Both prefix waits are $0.012$\,ms.
\looseness=-1

\subsection{Discussion}
\textbf{Ablation Studies.} Table~\ref{tab:ablation} ablates TR, OR, and RR on Fast-WAM-IDM using Table~\ref{tab:efficiency}'s 200-episode subset and evaluation setup; differences may reflect run-to-run variability. Full \system achieves $114.75$\,ms GPU time and the lowest O2A latency, $144.75$\,ms. TR and RR reduce computation: GPU time rises to $188.02$\,ms with 10-step denoising from noise without TR, and $188.99$\,ms with full DiT depth without RR. Removing OR restores real-conditioned execution without overlap, lowering GPU time to $108.45$\,ms but raising O2A latency to $165.62$\,ms. At matched step budgets without OR or RR, TR improves SR over fresh noise by 1.0 percentage point on both 3-step Cosmos Policy and 7-step Motus.
\looseness=-1

\textbf{Parameter Analysis.} Fig.~\ref{fig:success-tradeoff}(b) compares Default, Balanced, and Quality profiles. Quality achieves average SRs of $97.98\%$, $98.58\%$, and $85.55\%$, with O2A speedups of $1.20\times$, $2.06\times$, and $1.82\times$ on Cosmos, Fast-WAM-IDM, and Motus. Across the three profiles, RR fitting thresholds are $0.176/0.12/0.08$, $0.60/0.40/0.25$, and $0.40/0.30/0.20$, with accepted-prefix schedules $1+2/1+3/1+4$, video $2+4/2+5/2+6$ with action $5/6/7$, and $3+4/3+5/3+6$, respectively, under fixed OR thresholds and matching TR entry stages.
\looseness=-1

\textbf{Hardware Quantitative Analysis.} Fig.~\ref{fig:hardware-runtime} isolates stateful inference gains under matched CUDA Graph settings; without graphs, \system achieves $163.99$\,ms GPU time and $185.15$\,ms O2A latency. TR and OR each incur under $1$\,ms of CPU--GPU coordination per replan, whereas RR requires $4.557$\,ms and dominates D2D traffic at $1215.27$\,MiB, primarily for graph buffer staging within HBM.
\looseness=-1

We study how state reuse accelerates WAM inference and leave physical-robot deployment to future work, as robot-specific sensing and actuation delays are orthogonal to inference.

\FloatBarrier
\section{Conclusion}
\label{sec:conclusion}
In this paper, we first analyze state continuity and reuse challenges across three nested loops. We then present \system, a stateful inference framework that preserves and adapts states via Trajectory Remapping, Observation Rebinding, and Residual Rescaling. Across three WAM architectures, it achieves speedups of 1.47--3.05$\times$ in observation-to-action latency and 2.23--3.27$\times$ in GPU inference time per replan, with limited success rate degradation.
\looseness=-1

\section*{Content Generated by AI}
ChatGPT (OpenAI) assisted with drafting and polishing Sections III--VI.
The authors reviewed and verified all content.
\bibliographystyle{IEEEtran}
\bibliography{references}

@IEEEtranBSTCTL{IEEE:BSTcontrol,
  CTLuse_forced_etal = "yes",
  CTLmax_names_forced_etal = "6",
  CTLnames_show_etal = "1"
}

@inproceedings{kim2026cosmospolicy,
 author = {Kim, Moo Jin and Gao, Yihuai and Lin, Tsung-Yi and Lin, Yen-Chen and Ge, Yunhao and Lam, Grace and Liang, Percy and Song, Shuran and Liu, Ming-Yu and Finn, Chelsea and Gu, Jinwei},
 booktitle = {International Conference on Learning Representations},
 editor = {C. Vondrick and B. Hariharan and C. Raffel and L. Pinto and D. Yang and A. Faust},
 pages = {71531--71552},
 title = {Cosmos Policy: Fine-Tuning Video Models for Visuomotor Control and Planning},
 url = {https://proceedings.iclr.cc/paper_files/paper/2026/file/748becc400a57c0e31cfe6a2e7951467-Paper-Conference.pdf},
 volume = {2026},
 year = {2026}
}

@misc{yuan2026fastwam,
      title={Fast-WAM: Do World Action Models Need Test-time Future Imagination?}, 
      author={Tianyuan Yuan and Zibin Dong and Yicheng Liu and Hang Zhao},
      year={2026},
      eprint={2603.16666},
      archivePrefix={arXiv},
      primaryClass={cs.CV},
      url={https://arxiv.org/abs/2603.16666}, 
}

@InProceedings{bi2026motus,
    author    = {Bi, Hongzhe and Tan, Hengkai and Xie, Shenghao and Wang, Zeyuan and Huang, Shuhe and Liu, Haitian and Zhao, Ruowen and Feng, Yao and Xiang, Chendong and Rong, Yinze and Zhao, Hongyan and Liu, Hanyu and Su, Zhizhong and Ma, Lei and Su, Hang and Zhu, Jun},
    title     = {Motus: A Unified Latent Action World Model},
    booktitle = {Proceedings of the IEEE/CVF Conference on Computer Vision and Pattern Recognition (CVPR)},
    month     = {June},
    year      = {2026},
    pages     = {35101-35113}
}

@article{chi2025diffusionpolicy,
author = {Cheng Chi and Zhenjia Xu and Siyuan Feng and Eric Cousineau and Yilun Du and Benjamin Burchfiel and Russ Tedrake and Shuran Song},
title ={Diffusion policy: Visuomotor policy learning via action diffusion},

journal = {The International Journal of Robotics Research},
volume = {44},
number = {10-11},
pages = {1684-1704},
year = {2025},
doi = {10.1177/02783649241273668},

URL = {https://doi.org/10.1177/02783649241273668},
eprint = {https://doi.org/10.1177/02783649241273668}
}

@INPROCEEDINGS{black2025pi0, 
    AUTHOR    = {Kevin Black AND Noah Brown AND Danny Driess AND Adnan Esmail AND Michael Robert Equi AND Chelsea Finn AND Niccolo Fusai AND Lachy Groom AND Karol Hausman AND Brian Ichter AND Szymon Jakubczak AND Tim Jones AND Liyiming Ke AND Sergey Levine AND Adrian Li-Bell AND Mohith Mothukuri AND Suraj Nair AND Karl Pertsch AND Lucy Xiaoyang Shi AND Laura Smith AND James Tanner AND Quan Vuong AND Anna Walling AND Haohuan Wang AND Ury Zhilinsky}, 
    TITLE     = {{$\pi_0$: A Vision-Language-Action Flow Model for General Robot Control}}, 
    BOOKTITLE = {Proceedings of Robotics: Science and Systems}, 
    YEAR      = {2025}, 
    ADDRESS   = {LosAngeles, CA, USA}, 
    MONTH     = {June}, 
    DOI       = {10.15607/RSS.2025.XXI.010} 
}

@InProceedings{hou2024ditpolicy,
    author    = {Hou, Zhi and Zhang, Tianyi and Xiong, Yuwen and Duan, Haonan and Pu, Hengjun and Tong, Ronglei and Zhao, Chengyang and Zhu, Xizhou and Qiao, Yu and Dai, Jifeng and Chen, Yuntao},
    title     = {Dita: Scaling Diffusion Transformer for Generalist Vision-Language-Action Policy},
    booktitle = {Proceedings of the IEEE/CVF International Conference on Computer Vision (ICCV)},
    month     = {October},
    year      = {2025},
    pages     = {7686-7697}
}

@inproceedings{vaswani2017attention,
 author = {Vaswani, Ashish and Shazeer, Noam and Parmar, Niki and Uszkoreit, Jakob and Jones, Llion and Gomez, Aidan N and Kaiser, \L ukasz and Polosukhin, Illia},
 booktitle = {Advances in Neural Information Processing Systems},
 editor = {I. Guyon and U. Von Luxburg and S. Bengio and H. Wallach and R. Fergus and S. Vishwanathan and R. Garnett},
 pages = {},
 publisher = {Curran Associates, Inc.},
 title = {Attention is All you Need},
 url = {https://proceedings.neurips.cc/paper_files/paper/2017/file/3f5ee243547dee91fbd053c1c4a845aa-Paper.pdf},
 volume = {30},
 year = {2017}
}

@inproceedings{duan2025rtidp,
title = "Real-time Iteration Scheme for Diffusion Policy",
author = "Yufei Duan and Hang Yin and Danica Kragic",
year = "2025",
doi = "10.1109/IROS60139.2025.11247391",
language = "English",
pages = "11758--11764",
booktitle = "IROS 2025 - 2025 IEEE/RSJ International Conference on Intelligent Robots and Systems, Conference Proceedings",
publisher = "IEEE",
}

@InProceedings{li2026step,
  title = 	 {{STEP}: Warm-Started Visuomotor Policies with Spatiotemporal Consistency Prediction},
  author =       {Li, Jinhao and Cong, Yuxuan and Wang, Yingqiao and Xia, Hao and Huang, Shan and Zhang, Yijia and Xu, Ningyi and Dai, Guohao},
  booktitle = 	 {Proceedings of the 43rd International Conference on Machine Learning},
  pages = 	 {66501--66525},
  year = 	 {2026},
  editor = 	 {Zhang, Tong and Dudik, Miroslav and Jaggi, Martin and Agarwal, Alekh and Li, Sharon and Schuurmans, Dale and Zhu, Jerry and Berkenkamp, Felix and Dong, Hanze and Bietti, Alberto},
  volume = 	 {306},
  series = 	 {Proceedings of Machine Learning Research},
  month = 	 {06--11 Jul},
  publisher =    {PMLR},
  url = 	 {https://proceedings.mlr.press/v306/li26aa.html}
}

@inproceedings{black2025rtc,
 author = {Black, Kevin and Galliker, Manuel and Levine, Sergey},
 booktitle = {Advances in Neural Information Processing Systems},
 doi = {10.52202/085713-1122},
 editor = {D. Belgrave and C. Zhang and H. Lin and R. Pascanu and P. Koniusz and M. Ghassemi and N. Chen},
 pages = {33383--33407},
 publisher = {Curran Associates, Inc.},
 title = {Real-Time Execution of Action Chunking Flow Policies},
 url = {https://proceedings.neurips.cc/paper_files/paper/2025/file/300ccb2187dedd4edcc07f7e76d8e553-Paper-Conference.pdf},
 volume = {38, Main Conference},
 year = {2025}
}

@misc{jiang2026futurertc,
      title={FutureRTC: Real-Time Robot Execution with Anticipatory-Conditioned Action Chunking}, 
      author={Hai Jiang and Yixian Zou and Binbin Liang and Boqian Liu and Fanman Meng and Shuaicheng Liu},
      year={2026},
      eprint={2607.24008},
      archivePrefix={arXiv},
      primaryClass={cs.RO},
      url={https://arxiv.org/abs/2607.24008}, 
}

@InProceedings{liu2025teacache,
    author    = {Liu, Feng and Zhang, Shiwei and Wang, Xiaofeng and Wei, Yujie and Qiu, Haonan and Zhao, Yuzhong and Zhang, Yingya and Ye, Qixiang and Wan, Fang},
    title     = {Timestep Embedding Tells: It's Time to Cache for Video Diffusion Model},
    booktitle = {Proceedings of the IEEE/CVF Conference on Computer Vision and Pattern Recognition (CVPR)},
    month     = {June},
    year      = {2025},
    pages     = {7353-7363}
}

@inproceedings{bu2026dicache,
 author = {Bu, Jiazi and Ling, Pengyang and Zhou, Yujie and Wang, Yibin and Zang, Yuhang and Lin, Dahua and Wang, Jiaqi},
 booktitle = {International Conference on Learning Representations},
 editor = {C. Vondrick and B. Hariharan and C. Raffel and L. Pinto and D. Yang and A. Faust},
 pages = {73778--73803},
 title = {DiCache: Let Diffusion Model Determine Its Own Cache},
 url = {https://proceedings.iclr.cc/paper_files/paper/2026/file/78288ef33b18a351c3cd679dc9a15c8d-Paper-Conference.pdf},
 volume = {2026},
 year = {2026}
}

@misc{zhao2026c3ache,
      title={C$^3$ache: Accelerating World Action Models with Cross Inference Chunk Cache}, 
      author={Weisen Zhao and Lam Nguyen and Zhicong Lu and Yuzhang Shang},
      year={2026},
      eprint={2606.08962},
      archivePrefix={arXiv},
      primaryClass={cs.LG},
      url={https://arxiv.org/abs/2606.08962}, 
}

@manual{nvidia2026cudagraphs,
  title     = {{CUDA} Programming Guide: {CUDA Graphs}},
  author    = {{NVIDIA}},
  year      = {2026},
  note      = {{R}elease 13.2. Accessed September 25, 2026},
  url       = {https://docs.nvidia.com/cuda/archive/13.2.0/cuda-programming-guide/04-special-topics/cuda-graphs.html}
}

@inproceedings{liu2023libero,
 author = {Liu, Bo and Zhu, Yifeng and Gao, Chongkai and Feng, Yihao and Liu, Qiang and Zhu, Yuke and Stone, Peter},
 booktitle = {Advances in Neural Information Processing Systems},
 doi = {10.52202/075280-1939},
 editor = {A. Oh and T. Naumann and A. Globerson and K. Saenko and M. Hardt and S. Levine},
 pages = {44776--44791},
 publisher = {Curran Associates, Inc.},
 title = {LIBERO: Benchmarking Knowledge Transfer for Lifelong Robot Learning},
 url = {https://proceedings.neurips.cc/paper_files/paper/2023/file/8c3c666820ea055a77726d66fc7d447f-Paper-Datasets_and_Benchmarks.pdf},
 volume = {36},
 year = {2023}
}

@InProceedings{chen2026robotwin2,
  title = 	 {{R}obo{T}win 2.0: A Scalable Data Generator and Benchmark with Strong Domain Randomization for Robust Bimanual Robotic Manipulation},
  author =       {Chen, Tianxing and Chen, Zanxin and Chen, Baijun and Cai, Zijian and Liu, Yibin and Li, Zixuan and Liang, Qiwei and Lin, Xianliang and Ge, Yiheng and Gu, Zhenyu and Deng, Weiliang and Guo, Yubin and Nian, Tian and Xie, Xuanbing and Chen, Qiangyu and Su, Kailun and Xu, Tianling and Liu, Guodong and Hu, Mengkang and Gao, Huan-Ang and Wang, Kaixuan and Liang, Zhixuan and Qin, Yusen and Yang, Xiaokang and Luo, Ping and Mu, Yao},
  booktitle = 	 {Proceedings of the 43rd International Conference on Machine Learning},
  pages = 	 {13673--13699},
  year = 	 {2026},
  editor = 	 {Zhang, Tong and Dudik, Miroslav and Jaggi, Martin and Agarwal, Alekh and Li, Sharon and Schuurmans, Dale and Zhu, Jerry and Berkenkamp, Felix and Dong, Hanze and Bietti, Alberto},
  volume = 	 {306},
  series = 	 {Proceedings of Machine Learning Research},
  month = 	 {06--11 Jul},
  publisher =    {PMLR},
  url = 	 {https://proceedings.mlr.press/v306/chen26j.html}
}

@article{liang2025mot,
title={Mixture-of-Transformers: A Sparse and Scalable Architecture for Multi-Modal Foundation Models},
author={Weixin Liang and LILI YU and Liang Luo and Srini Iyer and Ning Dong and Chunting Zhou and Gargi Ghosh and Mike Lewis and Wen-tau Yih and Luke Zettlemoyer and Xi Victoria Lin},
journal={Transactions on Machine Learning Research},
issn={2835-8856},
year={2025},
url={https://openreview.net/forum?id=Nu6N69i8SB},
note={}
}

@misc{zhou2026tau0wm,
      title={$\tau_0$-WM: A Unified Video-Action World Model for Robotic Manipulation}, 
      author={Pengfei Zhou and Shengcong Chen and Di Chen and Jiaxu Wang and Rongjun Jin and Bingwen Zhu and Yike Pan and Songen Gu and Kuanning Wang and Shufeng Nan and Xingyu Qiu and Chenhao Qiu and Pu Yang and Yunuo Cai and Jianxiong Gao and Yifan Li and Yanwei Fu and Xiangyu Yue and Zhi Chen and Jianlan Luo},
      year={2026},
      eprint={2606.01027},
      archivePrefix={arXiv},
      primaryClass={cs.RO},
      url={https://arxiv.org/abs/2606.01027}, 
}

@INPROCEEDINGS{zhao2023act, 
    AUTHOR    = {Tony Z. Zhao AND Vikash Kumar AND Sergey Levine AND Chelsea Finn}, 
    TITLE     = {{Learning Fine-Grained Bimanual Manipulation with Low-Cost Hardware}}, 
    BOOKTITLE = {Proceedings of Robotics: Science and Systems}, 
    YEAR      = {2023}, 
    ADDRESS   = {Daegu, Republic of Korea}, 
    MONTH     = {July}, 
    DOI       = {10.15607/RSS.2023.XIX.016} 
}

@inproceedings{ho2020ddpm,
 author = {Ho, Jonathan and Jain, Ajay and Abbeel, Pieter},
 booktitle = {Advances in Neural Information Processing Systems},
 editor = {H. Larochelle and M. Ranzato and R. Hadsell and M.F. Balcan and H. Lin},
 pages = {6840--6851},
 publisher = {Curran Associates, Inc.},
 title = {Denoising Diffusion Probabilistic Models},
 url = {https://proceedings.neurips.cc/paper_files/paper/2020/file/4c5bcfec8584af0d967f1ab10179ca4b-Paper.pdf},
 volume = {33},
 year = {2020}
}

@inproceedings{karras2022edm,
 author = {Karras, Tero and Aittala, Miika and Aila, Timo and Laine, Samuli},
 booktitle = {Advances in Neural Information Processing Systems},
 doi = {10.52202/068431-1926},
 editor = {S. Koyejo and S. Mohamed and A. Agarwal and D. Belgrave and K. Cho and A. Oh},
 pages = {26565--26577},
 publisher = {Curran Associates, Inc.},
 title = {Elucidating the Design Space of Diffusion-Based Generative Models},
 url = {https://proceedings.neurips.cc/paper_files/paper/2022/file/a98846e9d9cc01cfb87eb694d946ce6b-Paper-Conference.pdf},
 volume = {35},
 year = {2022}
}

@inproceedings{lipman2023flowmatching,
title={Flow Matching for Generative Modeling},
author={Yaron Lipman and Ricky T. Q. Chen and Heli Ben-Hamu and Maximilian Nickel and Matthew Le},
booktitle={The Eleventh International Conference on Learning Representations },
year={2023},
url={https://openreview.net/forum?id=PqvMRDCJT9t}
}

@InProceedings{peebles2023dit,
    author    = {Peebles, William and Xie, Saining},
    title     = {Scalable Diffusion Models with Transformers},
    booktitle = {Proceedings of the IEEE/CVF International Conference on Computer Vision (ICCV)},
    month     = {October},
    year      = {2023},
    pages     = {4195-4205}
}

@INPROCEEDINGS{he2016resnet,
  author={He, Kaiming and Zhang, Xiangyu and Ren, Shaoqing and Sun, Jian},
  booktitle={2016 IEEE Conference on Computer Vision and Pattern Recognition (CVPR)}, 
  title={Deep Residual Learning for Image Recognition}, 
  year={2016},
  volume={},
  number={},
  pages={770-778},
  doi={10.1109/CVPR.2016.90}}

@misc{ma2026fasterwam,
      title={Faster-WAM: Do World Action Models Need Deep Action Modules?}, 
      author={Liheng Ma and Rui Heng Yang and Amin Abyaneh and George Z. Xue and Behnam Rahmati and Mateo Clemente and Ziwen Hu and Anlin Chen and Tongtong Cao and Zhanguang Zhang and Yingxue Zhang},
      year={2026},
      eprint={2608.02365},
      archivePrefix={arXiv},
      primaryClass={cs.AI},
      url={https://arxiv.org/abs/2608.02365}, 
}

@inproceedings{xu2025vlacache,
 author = {Xu, Siyu and Wang, Yunke and Xia, Chenghao and Zhu, Dihao and Huang, Tao and Xu, Chang},
 booktitle = {Advances in Neural Information Processing Systems},
 doi = {10.52202/085713-5484},
 editor = {D. Belgrave and C. Zhang and H. Lin and R. Pascanu and P. Koniusz and M. Ghassemi and N. Chen},
 pages = {164448--164473},
 publisher = {Curran Associates, Inc.},
 title = {VLA-Cache: Efficient Vision-Language-Action Manipulation via Adaptive Token Caching},
 url = {https://proceedings.neurips.cc/paper_files/paper/2025/file/f062da1973ac9ac61fc6d44dd7fa309f-Paper-Conference.pdf},
 volume = {38, Main Conference},
 year = {2025}
}

@inproceedings{ji2026bac,
 author = {Ji, Kangye and Meng, Yuan and Cui, Hanyun and Li, Ye and Zhou, Jianbo and Hua, Shengjia and Chen, Lei and Wang, Zhi},
 booktitle = {International Conference on Learning Representations},
 editor = {C. Vondrick and B. Hariharan and C. Raffel and L. Pinto and D. Yang and A. Faust},
 pages = {22805--22858},
 title = {Block-wise Adaptive Caching for Accelerating Diffusion Policy},
 url = {https://proceedings.iclr.cc/paper_files/paper/2026/file/2712b17bb58ea5b2b65c45857b024744-Paper-Conference.pdf},
 volume = {2026},
 year = {2026}
}
\end{document}